\documentclass{article} 
\usepackage{iclr2017_workshop,times}
\usepackage{hyperref}
\usepackage{url}
\usepackage{graphicx}
\usepackage{amssymb,amsmath}
\usepackage{microtype}
\usepackage{booktabs}
\title{
Towards ``AlphaChem'': 
Chemical Synthesis Planning with Tree Search and Deep Neural Network Policies}

\author{Marwin Segler$^{\clubsuit\nabla}$  \qquad  Mike Preu{\ss}$^\diamondsuit$ \\
$^\clubsuit$Institute of Organic Chemistry\\
$^\nabla$Center for Multiscale Theory and Computation\\
$^\diamondsuit$Department of Information Systems\\
Westf\"alische Wilhelms-Universit\"at M\"unster\\
\texttt{\{marwin.segler,mike.preuss\}@wwu.de} \\
\And
Mark P. Waller$^{\spadesuit\hbar}$\\
$^\spadesuit$Department of Physics\\
$^\hbar$International Center for Quantum\\ and Molecular Structures\\
Shanghai University\\
\texttt{waller@shu.edu.cn} \\
}

\DeclareMathOperator{\argmax}{arg\,max}

\begin{document}

\maketitle

\begin{abstract}
Retrosynthesis is a technique to plan the chemical synthesis of organic molecules, for example drugs, agro- and fine chemicals. In retrosynthesis, a search tree is built by analysing  molecules recursively and dissecting them into simpler molecular building blocks until one obtains a set of known building blocks. The search space is intractably large,  and it is difficult to determine the value of retrosynthetic positions. Here, we propose to model retrosynthesis as a Markov Decision Process. In combination with a Deep Neural Network policy learned from essentially the complete published knowledge of chemistry, Monte Carlo Tree Search (MCTS) can be used to evaluate positions. In exploratory studies, we demonstrate that MCTS with neural network policies outperforms the traditionally used best-first search with hand-coded heuristics.
\end{abstract}

\section{Introduction}
From medicines to materials, organic molecules are ubiquitous. They are indispensable for human well-being. To make them, chemists usually plan their syntheses with a ``working-backwards'' problem-solving technique called retrosynthesis [\cite{corey1989logic}]. In retrosynthesis, the target molecule is recursively analysed and dissected, until a set of known building blocks is obtained. Then, the chemist can go to the lab and execute the reaction steps starting with the buildings blocks. It is a formidable intellectual challenge that has strong analogies to puzzle games: There is a large number of rules, which are subsequently applied to the game state to give rise to a large, intractable search tree. While this tree is not as large as in Go, mainly due to the shallower depth of usually $\approx$10--20, retrosynthesis has a significant branching factor ($b\approx200$). An additional problem that arises in retrosynthesis, which is not very severe in Go or Chess, is to determine the applicable actions. Molecules are modelled as graphs, and the (re)actions are production rules on graphs which involve solving the NP-complete subgraph isomorphism problem. Thus, in a complete enumeration, for $N$ rules (for a non-toy system $N\approx10^4 - 10^5$), $N\times b^d$ NP-complete steps would have to be performed only to determine the legal actions!

Classically, automated retrosynthesis (computer-aided synthesis design) has been performed with very large expert systems based on tens of thousands of manually encoded reaction production rules and heuristics. Similar to other domains, despite of 40 years of active research these expert systems have not delivered convincing results yet [\cite{todd2005computer,szymkuc2016computer}]. However, retrosynthesis is in many aspects more similar to Go than to chess. No good heuristics to estimate state values have been found yet, which prohibits truncated search. The ``games'' thus have to be played out. 
Additionally, chemists rely on their intuition, which they master during long years of work and study, to prioritise which rules to apply when retroanalysing molecules. Analogous to master move prediction in Go (\cite{maddison2014move}), we showed recently that, instead of hand-coding, neural networks can learn the ``master chemist moves''. We extracted production rules from millions of reactions  (essentially the complete published knowledge of chemistry [\cite{reaxys}]) automatically, and then trained Neural Networks to predict the best transformation, or ``the best move'', to apply to a molecule for single step analyses [\cite{neural-symbolic}].
%
 In this work, we seek to transfer the other lessons learned from the last decade of computer Go, culminating in AlphaGo, to the full chemical synthesis planning problem by combining our neural networks as policies with Monte Carlo Tree Search (MCTS) [\cite{silver2016mastering,browne2012survey}]. 
\begin{figure}[t]
\begin{center}
\includegraphics[width=1.0\linewidth]{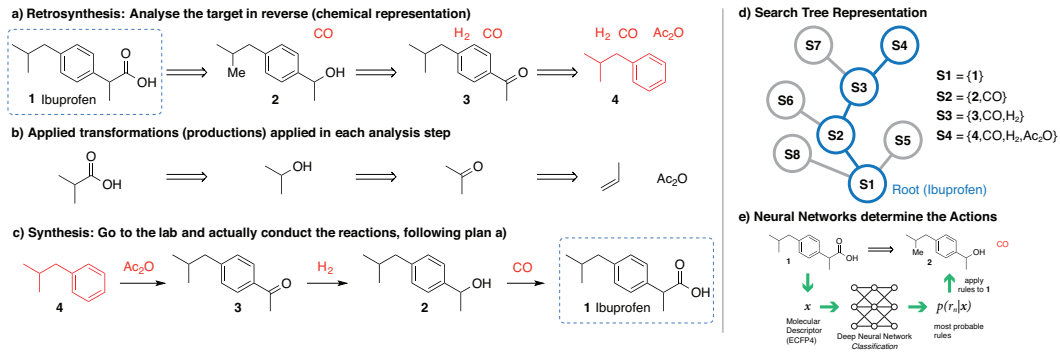}
\caption{a) The synthesis of the target (ibuprofen) is planned with retrosynthesis. The target is analyzed until a set of building blocks is found that we can buy or we already know how to make (molecules marked red). Double arrows indicate a ``reverse'' retrosynthetic analysis step. b) The transformation/production rules (actions) to analyse the molecules (state) c) After a synthesis route has been found, we can run the reactions in the lab to make the target. Single arrows indicate chemical reactions, chemicals over an arrow are added during the course of the reaction. d) Translation of the chemical representation (a) to a search tree. e) Neural networks are used as a policy to select the best actions and are trained on essentially the complete chemical literature since 1771.}
\label{fig:retro}
\end{center}
\end{figure}
The contribution of this work is as follows: 1) We provide the first analysis of retrosynthesis as a Markov Decision Process, 2) introduce MCTS to the synthesis planning problem to approximate the value of states and actions, and 3) show how Deep Neural Networks can be used as policies, to reduce the branching factor, and to approximate the subgraph isomorphism problem.

\section{Theory}

\subsection{Markov Decision Processes}

A Markov Decision Process (MDP) is defined by a state space $\mathcal{S}$, and an action space $\mathcal{A}(s)$, defining the legal actions in a given state $s \in \mathcal{S}$. Here, a state $s \in \mathcal{S}$ is a set of molecules $s=\{m_i,m_j,...\}$. The initial state $s_0 = \{m_0\}$ contains the target molecule $m_0$. Actions are productions on graphs. When applying a legal action $a_l$ to a state $s_i = \{m_a,m_b,m_c\}$, it will produce a new state, e.g. $s_j = \{m_d,m_b,m_c\}$. The transition function $T(s,a,s^\prime)=P(s^\prime | s,a)$ is set to be deterministic in this work for simplicity. The reward function $r(s)$ returns 1 if the state is terminal and solved (see below), -1 if the state is terminal and unsolved, and 0 otherwise. 
A policy $\pi(a|s)$ is a probability distribution over all actions $a_i \in \mathcal{A}(s)$, given state $s$. In this work, we use a deep highway network, introduced by \cite{srivastava2015training}, with the ELU nonlinearities by \cite{clevert2015fast} and a softmax output layer to parametrize $\pi(a|s)$. It predicts the probability distribution over all transformation rules.

\subsection{Chemistry}

Molecules are treated as molecular graphs, which are vertex-labeled and edge-labeled graphs $m=(A,B)$, with atoms A as vertices and bonds B as edges [\cite{andersen2014generic}]. Retrosynthetic disconnection rules are productions on graphs $p:\mathfrak{G} \rightarrow \mathfrak{M}$, where $\mathfrak{M}$ is the space of multisets of molecules in $\mathfrak{G}$. Most rules are unary and binary, that means they transform a molecule into one or two new graphs. Each action is a production applied to one of the molecules $m$ in state $s$. The rules are extracted automatically from the literature following established procedures [\cite{segler2017modelling}].
Given a set of building blocks $\mathfrak{R}$, the retrosynthetic analysis state $s_k$ is solved if $\forall m_i \in s_k: m_i \cong r_j \in \mathfrak{R}$, that is all molecules $m_i$ in $s_k$ are isomorphic to a building block $r_j \in \mathfrak{R}$.

\subsection{Monte Carlo Tree Search (MCTS)}
MCTS combines tree search with random sampling [\cite{browne2012survey}]. We use our policy network in three of the four MCTS steps: Selection, Expansion, and Rollout.
Within the search tree, we employ Eq. (1), a variation of the UCT function similar to AlphaGo's, to determine the best child $v^\prime$ of a node $v$, where $P(a)$ is the probability assigned to this action by the policy network, $N(v)$ the number of visits and $Q(v)$ the accumulated reward at node $v$, and $c$ the exploration constant set to 3. 
\begin{equation}
\underset{v^\prime \in \text{children of } v}{\argmax}\left(\frac{Q(v^\prime)}{N(v^\prime)} + c P(a) \frac{\sqrt{N(v)}}{1+N(v^\prime)}\right)
\end{equation} %
The tree policy is applied until a leaf node is found,  which is expanded by means of the policy network. %
Our policy network has a top1 accuracy of 0.62 and a top50 (of 17370) accuracy of 0.98. This allows to kill two birds with one stone: First, to reduce the branching factor, we only expand the top 50 most promising actions, and not all possible ones ($\approx 200$). 
Second, by evaluating only 50 actions, we have to solve the subgraph isomorphism problem, which determines if the corresponding rule can  be applied to a molecule and yields the next molecule(s), only 50 times instead of 17370 times for all rules. During rollout, we sample actions from the policy network until the problem is solved or a maximal depth of 20 is reached.

We compare our algorithm to the state of the art search method, which is Best-First Search (BFS) with hand-coded heuristics, as described by \cite{szymkuc2016computer}. In BFS, each branch is added to a priority queue, which is sorted by cost. Additionally, we perform BFS with the cost for a branch $b$ calculated with the policy network as $f(b) = \sum_{i=0}^{s \in b}(1-P(a_i|s_i))$.

\section{Results and Discussion}
For testing, we randomly selected 40 drug-like molecules generated with the neural chemical language model proposed  by \cite{segler2017generating}. These target molecules were analyzed with the different algorithms, to find a synthesis route to known building blocks. The algorithms received a budget of 2~h walltime to propose retrosynthetic paths for the 40 different targets. BFS was allowed to explore up to 9000 branches, MCTS performed 9000 simulations. All calculations were performed on a late 2013 MacBook Pro on CPU. 

BFS with heuristic cost is outperformed both in terms of time and accuracy by the BFS and MCTS with the deep neural network (DNN) policy. BFS+DNN is two times faster than MCTS+DNN, however, MCTS+DNN is able to solve more problems.

\begin{table}[htdp]
\caption{Experimental Results}
\begin{center}
\begin{tabular}{lllrrr}
\toprule
Entry & Search & Policy/Cost & solved/\% & timed out/\% & mean time per mol./s\\
\midrule
1 & BFS & Hand-coded Heuristic & 22.5 & 70 & 600 \\
2 & BFS & Highway Network & 87.5 & 0 & \textbf{29}\\
3 & \textbf{MCTS} & \textbf{Highway Network} & \textbf{95} & \textbf{0} & 68\\
\bottomrule
\end{tabular}
\end{center}
\label{tab:exp}
\end{table}%

\section{Conclusion}
In this work, we provided preliminary evidence that combining deep neural network policies with BFS and MCTS can also be used for planning in a highly challenging and important non-game domain, namely chemical synthesis planning. Our system outperforms the current state of the art, BFS with hand-coded heuristics. Larger scale experiments are currently ongoing. In future work, we seek to explore learning value functions to perform truncated search, learning better policies, and exploit transpositions. Another interesting avenue is to use more powerful, but slower algorithms to calculate values for transformations, for example our recently proposed model of chemical reasoning [\cite{segler2017modelling}]. We envision that our algorithm will become a valueable tool for synthetic chemists in industry and academia.

\subsubsection*{Acknowledgments}
M.S. and M.P.W. thank RELX Intellectual Properties for the reaction dataset and Deutsche Forschungsgemeinschaft (SFB858) for funding.

\bibliography{mcts}
\bibliographystyle{iclr2017_workshop}

\end{document}